\pdfoutput=1

\documentclass[11pt]{article}

\usepackage[final]{acl}

\usepackage{times}
\usepackage{latexsym}

\usepackage[T1]{fontenc}

\usepackage[utf8]{inputenc}

\usepackage{microtype}

\usepackage{inconsolata}

\usepackage{booktabs}
\usepackage{multirow}
\usepackage{subcaption}
\usepackage{graphicx}
\usepackage{soul}
\usepackage{color}

\newcommand{\hlc}[2][yellow]{{\sethlcolor{#1}\hl{#2}}}

\title{Evaluating Lexicon Incorporation for Depression Symptom Estimation}

\author{Kirill Milintsevich$^{1,2}$ \and Ga\"el Dias$^1$ \and  Kairit Sirts$^2$\\
        $^1$Normandie Univ, UNICAEN, ENSICAEN, CNRS, GREYC, France\\
        $^2$Institute of Computer Science, University of Tartu, Estonia\\
        \texttt{\{first\_name\}.\{last\_name\}@\{unicaen.fr$^1$|ut.ee$^2$\}}}

\begin{document}
\maketitle

\begin{abstract}
This paper explores the impact of incorporating sentiment, emotion, and domain-specific lexicons into a transformer-based model for depression symptom estimation. Lexicon information is added by marking the words in the input transcripts of patient-therapist conversations as well as in social media posts. Overall results show that the introduction of external knowledge within pre-trained language models can be beneficial for prediction performance, while different lexicons show distinct behaviours depending on the targeted task. Additionally, new state-of-the-art results are obtained for the estimation of depression level over patient-therapist interviews.
\end{abstract}

\section{Introduction}

Considerable interest has emerged in using natural language processing to unobtrusively infer one's mental health condition~\cite{chancellor2020methods}. A majority of studies have focused on predicting major depressive disorder (MDD) either as a symptom-based estimation~\cite{yadav-etal-2020-identifying,milintsevich2023towards} or a binary classification problem~\cite{burdissonode,xezonaki20_interspeech}. Both clinically motivated research initiatives and social media studies have emerged. In the latter case, Twitter~\cite{zhang2023phq}, Reddit~\cite{gupta-etal-2022-learning} and depression-related forums~\cite{yao2021extracting} have fostered attention. In the former case, recorded patient-therapist conversations are transcribed and associated with self-assessment depression questionnaires, such as PHQ-8~\cite{kroenke_phq-8_2009} or BDI~\cite{beck1988psychometric}.

\begin{table}[h]
\begin{center}
\small
\begin{tabular}{p{0.95\linewidth}}
\toprule
\textbf{Illustration of the lexicon-based input marking} \\
\midrule
a) i'm pretty much good because see by me being a bus operator you run into circumstances and situations you gotta remain calm and still remain professional at the same time \\
\midrule
b) i'm \hlc[yellow]{@ pretty @} much \hlc[yellow]{@ good @} because see by me being a bus operator you run into circumstances and situations you gotta remain \hlc[yellow]{@ calm @} and still remain professional at the same time\\
\midrule
c) i'm \hlc[green]{@ pretty @} much \hlc[green]{@ good @} because see by me being a bus operator you run into circumstances and situations you gotta remain \hlc[green]{@ calm @} and still remain \hlc[green]{@ professional @} at the same \hlc[green]{@ time @}\\
\bottomrule
\end{tabular}
\caption{Example of input marking. Text a) is the original text without markings, \hlc[yellow]{b)} and \hlc[green]{c)} show text with terms from \hlc[yellow]{AFINN} and \hlc[green]{NRC} lexicons.}
\label{tab:marking}
\end{center}
\end{table}

The DAIC-WOZ dataset~\cite{gratch-etal-2014-distress} has mostly been studied within the context of clinical research. Different works have been proposed to automatically infer depression level on this dataset: multi-modal~\cite{qureshi2019multitask,wei2023multimodal} and text-based architectures~\cite{li2023detecting,agarwal2022agent}. The PRIMATE dataset~~\cite{gupta-etal-2022-learning} has also received recent attention within the context of early symptom prediction on social media posts. The most comprehensive work on this dataset is proposed by~\citet{zhang2023phq}, which defines a context- and PHQ-aware transformer-based architecture.    

People with MDD have shown increased use of negative emotional words and decreased use of positive emotional words ~\cite{rude2004language,savekar2023structural}.
In this line, \citet{xezonaki20_interspeech} and \citet{qureshi2020improving} used feature-level and task fusion of emotion and sentiment knowledge and showed improved performance for depression estimation. However, these works, along with other studies on social media mental health data~\cite{zhang2023emotion}, have used pre-transformer era neural architectures.
Recent state-of-the-art approaches that rely on transformer-based pre-trained language models (PLMs) have not explored external knowledge fusion~\cite{milintsevich2023towards}. 

In this paper, we investigate whether pre-trained language models could benefit from the introduction of emotional, sentimental, and domain-specific external knowledge from the lexicons: AFINN~\cite{nielsen2011new}, NRC~\cite{Mohammad13} and SDD~\cite{yazdavar2017semi}. Introducing this external knowledge into a transformer-based model is feature-level and is achieved by modifying the input with specific markers that highlight spans of text, as shown in Table~\ref{tab:marking}, inspired by the works of \citet{wang-etal-2021-k} and \citet{zhou-chen-2022-improved}. This approach does not require any modification to the model's architecture, such as changing attention mechanism~\cite{li-etal-2021-improving-bert,wang2022paying} or adding new layers~\cite{bai-etal-2022-enhancing}; it also keeps the model's vocabulary unchanged unlike \citet{zhong-chen-2021-frustratingly}.

Results on the DAIC-WOZ dataset show that the performance of transformer-based models is impacted by the added lexicon information (especially sentiment), and new state-of-the-art values can be obtained from the combination of the three lexicons. However, such results are less expressive for the PRIMATE dataset, with slight improvements induced by the introduction of external information. Overall, the improvement in predicting particular symptoms evidences that lexicon information can be helpful, provided that its content closely corresponds to the targeted task.

\section{Methodology}
\label{sec:methodology}

\begin{table}[t]
\begin{center}
\small
\begin{tabular}{lcrrr}
\toprule
Lexicon                & PHQ-8     & Train & Dev  & Test \\
\midrule
\multirow{2}{*}{AFINN} & $\geq 10$ & 8.4 & 7.6 & 8.0 \\
                       & $<10$     & 8.2 & 7.6 & 7.9 \\
\midrule
\multirow{2}{*}{NRC}   & $\geq 10$ & 7.6 & {}\textsuperscript{\textdagger}6.8 & {}\textsuperscript{\textdagger}7.1 \\
                       & $<10$     & 7.7 & {}\textsuperscript{\textdagger}7.6 & {}\textsuperscript{\textdagger}7.6 \\
\midrule
\multirow{2}{*}{SDD}   & $\geq 10$ & {}\textsuperscript{\textdagger}0.6 & 0.4 & 0.5 \\
                       & $<10$     & {}\textsuperscript{\textdagger}0.4 & 0.3 & 0.4 \\
\bottomrule
\end{tabular}
\caption{Proportion of marked words for each lexicon over the DAIC-WOZ. Reported values are in percentage. {\textdagger} shows if the difference between the depressed and non-depressed populations is statistically significant.}
\label{tab:phq_lex}
\end{center}
\end{table}

\paragraph{Data.}
In this work, we use two depression datasets: DAIC-WOZ~\citep{gratch-etal-2014-distress} and PRIMATE~\citep{gupta-etal-2022-learning}. 
The DAIC-WOZ dataset contains 189 clinical interviews in a dialogue format. Each interview has two actors: a human-controlled virtual therapist and a participant. The dataset is distributed in pre-determined splits, such that 107 interviews are used for training, 35 for validation, and 47 for testing. 
Each interview in the dataset is accompanied with a PHQ-8 assessment, which consists of eight questions inquiring about symptoms. Each question is scored from 0 to 3 on a Likert scale, and the total PHQ score ranging from 0 to 24 is the sum of the eight symptom scores. According to a standard cutoff score of 10, the interviews can be divided into diagnostic classes, where subjects with PHQ-8 total score $<10$ are considered non-depressed, and those with score $\geq10$ are categorized as depressed. The eight listed symptoms are: \texttt{LOI} (lack of interest), \texttt{DEP} (feeling down), \texttt{SLE} (sleeping disorder), \texttt{ENE} (lack of energy), \texttt{EAT} (eating disorder), \texttt{LSE} (low self-esteem), \texttt{CON} (concentration problem), \texttt{MOV} (hyper/lower activity).

The PRIMATE dataset is based on Reddit posts from depression-related communities, or subreddits, in which people describe their health conditions. A total of 2003 posts were manually annotated with binary labels for each individual symptom from the PHQ-9~\citep{kroenke2001phq}, each label signifying whether the corresponding symptom is discussed in the post or not. PHQ-9 has the same first eight symptoms as PHQ-8 and one additional \texttt{SUI} (suicidal thoughts). The data was labeled by five crowd workers and verified by a mental health professional. The dataset is not pre-split into the train, validation, and test sets, so we randomly take 1601, 201, and 201 posts for each split accordingly.

\begin{figure}[t]
    \centering
    \includegraphics[width=\columnwidth]{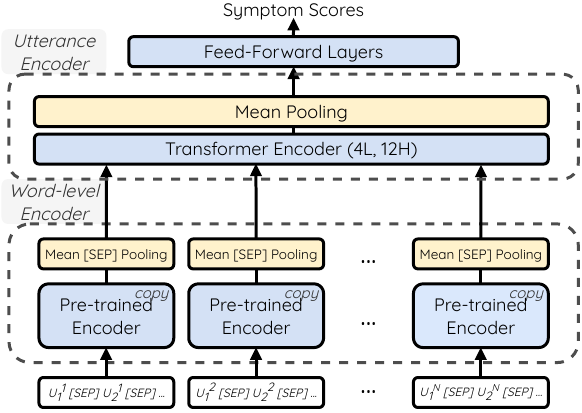}
    \caption{Overview of the model architecture. $U_i^N$ stands for $i$-th utterance of $N$-th input. \emph{Symptom Scores} are $||L||$ real numbers, where $||L||$ is the number of symptoms to predict.}
    \label{fig:model}
\end{figure}

\begin{table*}[t]
\setlength{\tabcolsep}{4pt}
\begin{center}
    \small
    \begin{tabular}{llllllllll}
    \toprule
        \textbf{Model} & \textbf{LOI} & \textbf{DEP} & \textbf{SLE} & \textbf{ENE} & \textbf{EAT} & \textbf{LSE} & \textbf{CON} & \textbf{MOV} & \textbf{PHQ-8} \\
    \midrule
        \textsc{BERT} & $0.56_{\pm .05}$ & \boldmath$0.63_{\pm .02}$ & $0.77_{\pm .05}$ & $0.87_{\pm .04}$ & \boldmath$0.81_{\pm .03}$ & $0.78_{\pm .06}$ & $0.74_{\pm .01}$ & $0.34_{\pm .01}$ & $4.38_{\pm .21}$ \\ 
    \midrule
        \textsc{+SDD} & $0.70_{\pm .02}$ & $0.88_{\pm .05}$ & $0.94_{\pm .05}$ & $0.94_{\pm .04}$ & $1.00_{\pm .07}$ & $0.97_{\pm .04}$ & $0.87_{\pm .02}$ & $0.34_{\pm .00}$ & $5.60_{\pm .18}$ \\ 
        \textsc{+AFINN} & \boldmath$0.50_{\pm .03}$ & $0.70_{\pm .03}$ & $0.79_{\pm .03}$ & $0.81_{\pm .04}$ & $0.85_{\pm .03}$ & $0.72_{\pm .02}$ & $0.77_{\pm .02}$ & $0.34_{\pm .00}$ & $4.56_{\pm .22}$ \\ 
        \textsc{+NRC} & \boldmath$0.50_{\pm .03}$ & $0.66_{\pm .05}$ & \boldmath$0.73_{\pm .05}$ & $0.77_{\pm .03}$ & \boldmath$0.81_{\pm .05}$ & $0.71_{\pm .07}$ & \boldmath$0.73_{\pm .05}$ & $0.34_{\pm .00}$ & \boldmath$4.31_{\pm .18}$ \\ 
        \textsc{+ALL} & \boldmath$0.50_{\pm .04}$ & $0.69_{\pm .03}$ & $0.81_{\pm .12}$ & \boldmath$0.74_{\pm .06}$ & \boldmath$0.81_{\pm .07}$ & \boldmath$0.69_{\pm .05}$ & $0.74_{\pm .03}$ & $0.34_{\pm .00}$ & $4.56_{\pm .42}$ \\
    \midrule
    \midrule
        \textsc{MeBERT} & $0.59_{\pm .02}$ & $0.64_{\pm .06}$ & $0.91_{\pm .05}$ & $0.92_{\pm .04}$ & $0.89_{\pm .04}$ & $0.71_{\pm .02}$ & $0.71_{\pm .04}$ & $0.35_{\pm .01}$ & $4.71_{\pm .23}$ \\   
    \midrule
        \textsc{+SDD} & $0.69_{\pm .07}$ & $0.72_{\pm .08}$ & $0.89_{\pm .07}$ & $0.92_{\pm .02}$ & $0.93_{\pm .07}$ & $0.85_{\pm .07}$ & $0.78_{\pm .06}$ & $0.34_{\pm .00}$ & $5.07_{\pm .38}$ \\
        \textsc{+AFINN} & $0.48_{\pm .04}$ & $0.62_{\pm .02}$ & $0.71_{\pm .05}$ & $0.78_{\pm .04}$ & $0.79_{\pm .03}$ & $0.70_{\pm .03}$ & $0.74_{\pm .03}$ & $0.34_{\pm .00}$ & $4.27_{\pm .22}$ \\
        \textsc{+NRC} & $0.60_{\pm .05}$ & $0.68_{\pm .03}$ & $0.71_{\pm .05}$ & $0.78_{\pm .04}$ & $0.80_{\pm .08}$ & $0.74_{\pm .02}$ & $0.71_{\pm .05}$ & $0.34_{\pm .00}$ & $4.35_{\pm .26}$ \\
        \textsc{+ALL} & \boldmath$0.44_{\pm .06}$ & \boldmath$0.55_{\pm .04}$ & \boldmath$0.63_{\pm .06}$ & \boldmath$0.72_{\pm .07}$ & \boldmath$0.69_{\pm .03}$ & \boldmath$0.67_{\pm .04}$ & \boldmath$0.67_{\pm .03}$ & $0.34_{\pm .00}$ & \boldmath$3.59_{\pm .31}$ \\
    \midrule
    \midrule
        \textsc{Sota} & $0.53_{\pm .05}$ & \boldmath$0.55_{\pm .03}$ & $0.75_{\pm .07}$ & \boldmath$0.64_{\pm .03}$ & $0.81_{\pm .05}$ & \boldmath$0.62_{\pm .02}$ & $0.83_{\pm .04}$ & $0.44_{\pm .02}$ & $3.78_{\pm .13}$ \\
    \bottomrule
    \end{tabular}
    \caption{Results for the DAIC-WOZ test set. The mean MAE and standard deviation are reported for five runs. The best MAE for each symptom is \textbf{in bold}. \textsc{Sota} means current state-of-the-art results in the literature \cite{milintsevich2023towards}.}
    \label{tab:results}
\end{center}
\end{table*}

\paragraph{Model architecture.}

To encode the interview transcripts, we adopt the hierarchical model from \cite{milintsevich2023towards}. In their model, the interview is first split utterance-by-utterance, with each utterance processed by a word-level encoder. All utterance representations are then concatenated into one sequence, later processed by an utterance-level encoder. In the end, the classification head produces a real number in the range from $0$ to $3$ for each symptom. Several changes are made to the original architecture to gain training efficiency. First, the BiLSTM utterance-level encoder is replaced with a randomly initialized 4-layer 12-head transformer encoder. Second, we change the way the input data is represented. In the original model, each utterance of the interview is encoded separately by a word-level encoder. This is far from optimal since most of the utterances are short (<10 tokens), thus, a lot of computation is wasted on padding tokens. Instead, the utterances are concatenated into one input text separated by the \texttt{[SEP]} special token. This way, the number of passes through the encoder is reduced from the number of utterances $K$ to $\bar{K}$, defined as in Equation~\ref{eq:inputs}, where $|U_i|$ is the number of tokens in an utterance and $m$ is the maximum input length of the word-level encoder.
\begin{equation}
    \bar{K} = \left\lceil \frac{\sum \left(|U_i| + 1\right)}{m} \right\rceil
    \label{eq:inputs}
\end{equation}
\noindent In practice, it reduces the number of word-level encoder passes by ${\sim}40$ times for each input. After, we perform the \emph{Mean \texttt{[SEP]} pooling} on the tokens representing each utterance to get the final utterance representation. The overview of the model architecture is presented in Figure~\ref{fig:model}.

\begin{figure*}[t]
  \includegraphics[width=0.5\linewidth]{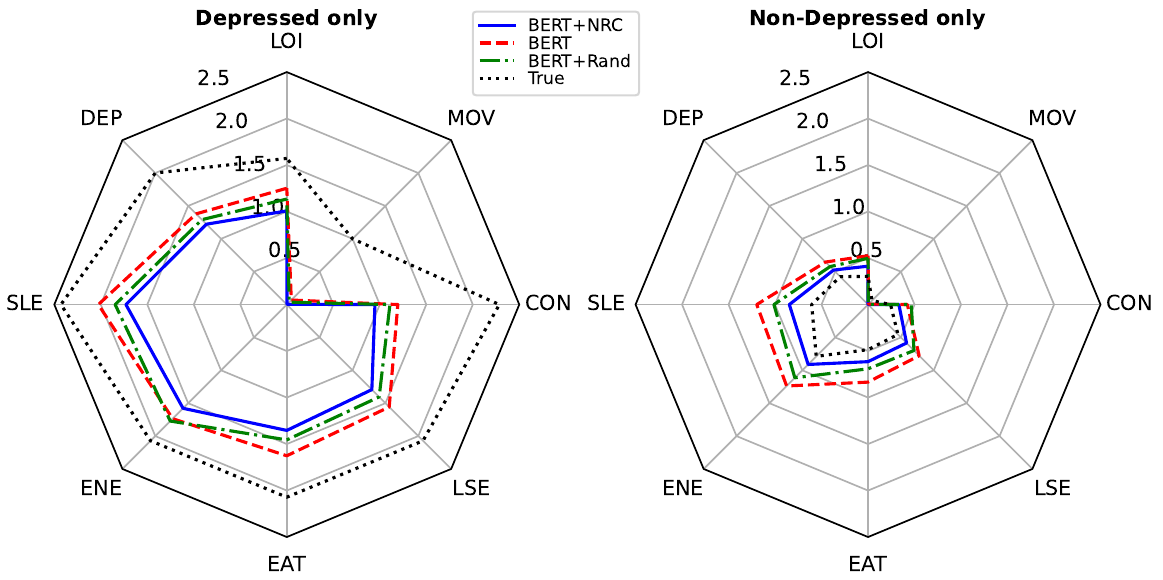} \hfill
  \includegraphics[width=0.5\linewidth]{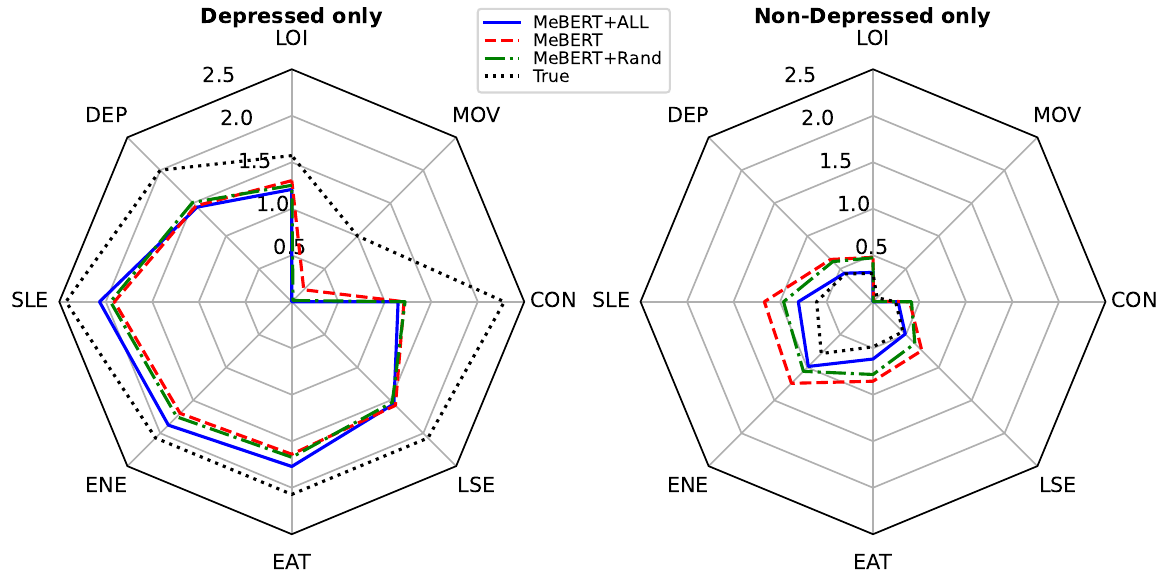}
  \caption {Average predicted values for depressed and non-depressed patients of the DAIC-WOZ test set.}
  \label{fig:radar_plots}
\end{figure*}

\begin{table*}[t]
\setlength{\tabcolsep}{4pt}
\begin{center}
    \small
    \begin{tabular}{llllllllll}
    \toprule
        \textbf{Model} & \textbf{LOI} & \textbf{DEP} & \textbf{SLE} & \textbf{ENE} & \textbf{EAT} & \textbf{LSE} & \textbf{CON} & \textbf{MOV} & \textbf{SUI} \\
    \midrule
        \textsc{BERT} & \boldmath$0.59_{\pm .03}$ & \boldmath$0.65_{\pm .03}$ & $0.81_{\pm .01}$ & $0.62_{\pm .02}$ & $0.75_{\pm .06}$ & $0.60_{\pm .02}$ & \boldmath$0.65_{\pm .01}$ & $0.81_{\pm .01}$ & $0.82_{\pm .01}$ \\ 
    \midrule
        \textsc{+SDD} & $0.58_{\pm .03}$ & $0.62_{\pm .02}$ & $0.81_{\pm .01}$ & \boldmath$0.64_{\pm .03}$ & $0.74_{\pm .03}$ & \boldmath$0.63_{\pm .03}$ & $0.63_{\pm .03}$ & \boldmath$0.82_{\pm .02}$ & $0.82_{\pm .01}$ \\ 
        \textsc{+AFINN} & $0.57_{\pm .03}$ & $0.60_{\pm .03}$ & $0.80_{\pm .02}$ & $0.62_{\pm .02}$ & $0.76_{\pm .02}$ & $0.59_{\pm .03}$ & $0.64_{\pm .01}$ & $0.81_{\pm .02}$ & \boldmath$0.83_{\pm .01}$ \\ 
        \textsc{+NRC} & $0.55_{\pm .04}$ & $0.62_{\pm .04}$ & \boldmath$0.82_{\pm .01}$ & $0.60_{\pm .02}$ & $0.79_{\pm .04}$ & $0.59_{\pm .03}$ & $0.61_{\pm .04}$ & $0.80_{\pm .01}$ & $0.82_{\pm .02}$ \\ 
        \textsc{+ALL} & $0.56_{\pm .05}$ & $0.63_{\pm .02}$ & $0.79_{\pm .02}$ & $0.61_{\pm .02}$ & \boldmath$0.80_{\pm .02}$ & $0.58_{\pm .03}$ & $0.61_{\pm .01}$ & \boldmath$0.82_{\pm .01}$ & $0.82_{\pm .02}$ \\ 
    \midrule
    \midrule
        \textsc{MeBERT} & \boldmath$0.58_{\pm .03}$ & $0.58_{\pm .02}$ & $0.82_{\pm .02}$ & $0.62_{\pm .01}$ & $0.78_{\pm .03}$ & $0.60_{\pm .04}$ & $0.62_{\pm .03}$ & \boldmath$0.82_{\pm .01}$ & $0.84_{\pm .01}$ \\ 
    \midrule
        \textsc{+SDD} & $0.53_{\pm .04}$ & \boldmath$0.60_{\pm .02}$ & \boldmath$0.83_{\pm .01}$ & $0.62_{\pm .02}$ & $0.79_{\pm .01}$ & $0.60_{\pm .02}$ & $0.61_{\pm .03}$ & $0.81_{\pm .02}$ & \boldmath$0.86_{\pm .01}$ \\ 
        \textsc{+AFINN} & $0.57_{\pm .03}$ & $0.55_{\pm .04}$ & \boldmath$0.83_{\pm .01}$ & $0.62_{\pm .02}$ & $0.79_{\pm .01}$ & \boldmath$0.63_{\pm .02}$ & $0.58_{\pm .02}$ & $0.81_{\pm .02}$ & $0.85_{\pm .02}$ \\ 
        \textsc{+NRC} & $0.57_{\pm .03}$ & $0.58_{\pm .03}$ & $0.82_{\pm .02}$ & \boldmath$0.63_{\pm .03}$ & $0.79_{\pm .02}$ & \boldmath$0.63_{\pm .01}$ & $0.61_{\pm .03}$ & $0.80_{\pm .02}$ & $0.85_{\pm .01}$ \\ 
        \textsc{+ALL} & $0.56_{\pm .03}$ & $0.59_{\pm .04}$ & $0.80_{\pm .02}$ & $0.62_{\pm .02}$ & \boldmath$0.80_{\pm .02}$ & $0.61_{\pm .01}$ & \boldmath$0.63_{\pm .02}$ & \boldmath$0.82_{\pm .02}$ & $0.84_{\pm .01}$ \\ 

    \bottomrule
    \end{tabular}
    \caption{Results for the PRIMATE test set. The mean macro-F1 score is reported for five runs. The best macro-F1 for each symptom is \textbf{in bold}. As standard splits are not provided, we cannot present \textsc{Sota} results. As standard splits are not provided, we cannot present \textsc{Sota} results.}
    \label{tab:results_primate}
\end{center}
\end{table*}

\paragraph{Lexicons.}
To incorporate the external knowledge into the model, we use three lexicons: AFINN~\cite{nielsen2011new}, NRC~\cite{Mohammad13}, and SDD~\cite{yazdavar2017semi}. AFINN is a sentiment lexicon that includes a list of 2,477 terms manually rated for the sentiment valence with a value between $-5$ (negative) and $+5$ (positive). \citet{nielsen2011new} used Twitter postings together with different word lists as a source for the lexicon. NRC is a word-emotion association lexicon that is a list of 14,182 words and their associations with eight basic emotions (anger, fear, anticipation, trust, surprise, sadness, joy, and disgust) and two sentiments (negative and positive). \citet{Mohammad13} compiled terms from Macquarie Thesaurus~\citep{bernard1986macquarrie}, WordNet Affect Lexicon~\citep{strapparava-valitutti-2004-wordnet}, and General Inquirer~\cite{stone1966general} and labeled them with the help of crowd-sourced workers. SDD is a part of the Social-media Depression Detector and is a lexicon of more than 1,620 depression-related words and phrases created in collaboration with a psychologist clinician. 

\paragraph{Input marking.}
In particular, we employ the technique proposed by Zhou and Chen~\cite{zhou-chen-2022-improved} to identify and annotate the lexicon words in the input text. It involves marking a lexicon word using the "\texttt{@}" token on either side (see Table~\ref{tab:marking} for examples). We chose the "\texttt{@}" token for marking since it is not present in the data but included in the model's vocabulary. This way, the pre-trained model's architecture remains unchanged\footnote{Typed marking strategies that include emotion and sentiment values have also been tested and provided no additional insights compared to the simple input marking.}. The proportion of marked words within the DAIC-WOZ is illustrated in Table \ref{tab:phq_lex}, where the statistical test is Student's t-test with p-value $<0.05$. 

\paragraph{Experimental setup.}
We used two pre-trained models in the word-level encoder of our architecture: BERT-Base model~\citep{DBLP:journals/corr/abs-1810-04805} and MentalBERT~\citep{ji2022mentalbert}. We refer to them as \textbf{BERT} and \textbf{MeBERT} further on. Both models share the same architecture; however, BERT was pre-trained on general domain data, while MeBERT used mental health-related data, mostly based on Reddit.
Each model is finetuned with the same hyperparameters (mostly following \citealp{mosbach2020stability}) and different input markings. For example, the BERT+SDD model uses BERT as a pre-trained model and SDD lexicon for input marking. +ALL models use a union of all three lexicons. All models are trained with a mini-batch size of 16, PyTorch realization of AdamW optimizer~\citep{loshchilov2017decoupled} with a learning rate of $2 \cdot 10^{-5}$ and linear scheduler with a warm-up ratio of $0.1$. For the word-level PLMs, only their attention layers are finetuned. The utterance-level encoder is randomly initialized based on the transformer encoder architecture with the following hyperparameters: $4$ layers, $12$ attention heads, hidden dimensions of encoder and pooler layers of $768$, intermediate hidden dimension of $1536$. The rest of the hyperparameters follow the default \texttt{BertConfig} from the HuggingFace Transformers library~\citep{wolf-etal-2020-transformers}.
For the DAIC-WOZ dataset, results are evaluated with micro-averaged mean absolute error (MAE). Symptom-based errors are calculated for each symptom individually. PHQ-8 score is obtained by summing the eight symptom scores, and MAE for PHQ-8 is calculated on this summation. We evaluate results on the PRIMATE dataset with a macro-averaged F1 score.

\section{Results and Discussion}

Table~\ref{tab:results} shows the results for the DAIC-WOZ test set.
For the BERT model, the lexicon-based input marking brings slight overall improvement when AFINN or NRC lexicons are introduced. Most notably, the NRC input marking shows improved or equal MAE for all symptom scores except \texttt{DEP}. The combination of all lexicons is marginally beneficial overall, and results have deteriorated with the exclusive introduction of the SDD lexicon. On the other hand, for the MeBERT model, the combination of all the lexicons produces the best results overall, both symptom-wise and for the global PHQ-8 score. Furthermore, both AFINN and NRC lexicons improve the prediction for the MeBERT model, similar to the BERT model. Also, when only the SDD lexicon is used for input marking, the model shows worse performance than the baseline setting.

Figure~\ref{fig:radar_plots} depicts a more detailed overview of the best-performing models: BERT+NRC and MeBERT+ALL. Additionally, we finetune the +Rand version of both BERT and MeBERT to verify if the improvement comes only from the input marking by randomly marking 8\% of the words in each interview. From the results, the improvement for the BERT+NRC model comes from the non-depressed population.
MeBERT+All model, however, improves for both depressed and non-depressed populations and is less sensitive to the marking bias.
Interestingly, +Rand models show some improvement for the non-depressed population, suggesting that input markings alone act as a regularizer.



Table~\ref{tab:results_primate} shows the results for the PRIMATE test set. Contrary to the results from Table~\ref{tab:results}, introducing external knowledge does not clearly improve performances. The models that use the lexicon input marking show signs of improvement for some symptoms, but it is largely inconsistent. Unlike for the DAIC-WOZ, the SDD-based input marking provides the best F1 score for three symptoms, both for BERT and MentalBERT models, while the benefits of AFINN and NRC are limited or absent and spread over symptoms.

The results from the DAIC-WOZ show that PLMs can indeed benefit from the introduction of external knowledge about the sentiment and emotional value of the words. Surprisingly, the introduction of the depression-specific lexicon had the opposite effect. We hypothesize that two reasons could cause it. First, as seen in Table~\ref{tab:phq_lex}, SDD covers less than 0.5\% of words in the interview, almost 15 times less than AFINN and NRC. Thus, the introduced signal might be too weak for the model to learn. Second, the SDD lexicon was based on Twitter data, while DAIC-WOZ contains transcripts of real conversations. From our observations, the people describe their problems more explicitly in their social media posts. At the same time, DAIC-WOZ conversations are more generally themed, and the PHQ-8 scores are based on the person's self-assessment test rather than the conversations themselves. This brings us back to the conceptual difference between the DAIC-WOZ and PRIMATE datasets. While the first one aims at establishing the link between the underlying person's mental condition and their speech, the latter one sets a goal of detecting whether a particular symptom is mentioned in the text. In addition, the PRIMATE dataset is annotated by layman crowd workers, and the labels are not consistent and contain inevitable mistakes~\cite{milintsevich-etal-2024-model}. This might explain the reason behind the greater impact of the AFINN and NRC lexicons for modeling the DAIC-WOZ dataset.

\section{Conclusion}
This paper targets lexicon incorporation in transformer-based models for symptom-based depression estimation. The external information is supplied through a marking strategy, which avoids any modification to the model's architecture. The set of endeavoured experiments shows that introducing sentimental, emotional and/or domain-specific lexicons can correlate with overall performance improvement if adapted to the targeted task\footnote{Source code is available here: \url{https://github.com/501Good/dialogue-classifier}.}.

\section*{Limitations}
The main limitation in automated clinical mental health assessment with natural language processing is the difficulty of acquiring and accessing large quantities of data. DAIC-WOZ and PRIMATE are rare exceptions as it is publicly available and clinically verified. However, DAIC-WOZ, in particular, suffers from a small number of data points that makes it hard to train and validate hypotheses, as both validation and test sets are particularly small. As a consequence, this piece of research requires further validation on a larger body of clinical data.

\section*{Ethical Considerations}
We acknowledge the potential ethical aspects of the work that studies the methods to unobtrusively detect someone’s mental health status. Here, we are using publicly available datasets collected for research purposes. Also, the lexicons we use are publicly available and have not been composed based on private confidential material. If such a system that could predict the presence of depression symptoms based on actual clinical interviews would be deployed in practice, it would require the informed consent of all participants involved as well as the understanding of the validity boundaries of such systems, meaning that the predictions of such systems cannot replace the assessment of trained clinicians, but rather assist them in their activities.

\section*{Acknowledgements}

This research was supported by the Estonian Research Council Grant PSG721 and the FHU A$^2$M$^2$P project funded by the G4 University Hospitals of Amiens, Caen, Lille and Rouen (France). The calculations for model’s training and inference were carried out in the High Performance Computing Center of the University of Tartu~\citep{https://doi.org/10.23673/ph6n-0144}.

\bibliography{custom}

\end{document}